\documentclass[letterpaper, 10 pt, conference]{ieeeconf}

\IEEEoverridecommandlockouts

\usepackage{times}
\usepackage{hyperref}
\usepackage{todonotes}
\usepackage[numbers]{natbib}
\usepackage{multicol}
\usepackage{graphics,graphicx,caption,float,subcaption,booktabs,xcolor,multirow,array,color,ifthen,tabu,colortbl,dblfloatfix,url,xparse,mathtools,patchcmd,algorithm,algpseudocode,xspace,nicefrac,microtype,amsmath,amsfonts,bm,ragged2e,tikz,stackengine,etoolbox,xpatch,enumerate,xstring,setspace,tabularx,makecell,changepage,cuted,titlesec,wrapfig,tcolorbox, sidecap,comment, bbding, placeins}
\usepackage{caption}
\usepackage{gensymb,comment}

\title{\LARGE \bf
IFG: Internet-Scale Guidance for Functional Grasping Generation
}

\author{Ray Muxin Liu*$\qquad$Mingxuan Li*$\qquad$Kenneth Shaw$\qquad$Deepak Pathak}
\newcommand{\our} {IFG\xspace}

\begin{document}

\makeatletter
\let\@oldmaketitle\@maketitle
\renewcommand{\@maketitle}{\@oldmaketitle
  \vspace{0.1in}
  \includegraphics[width=1\linewidth]{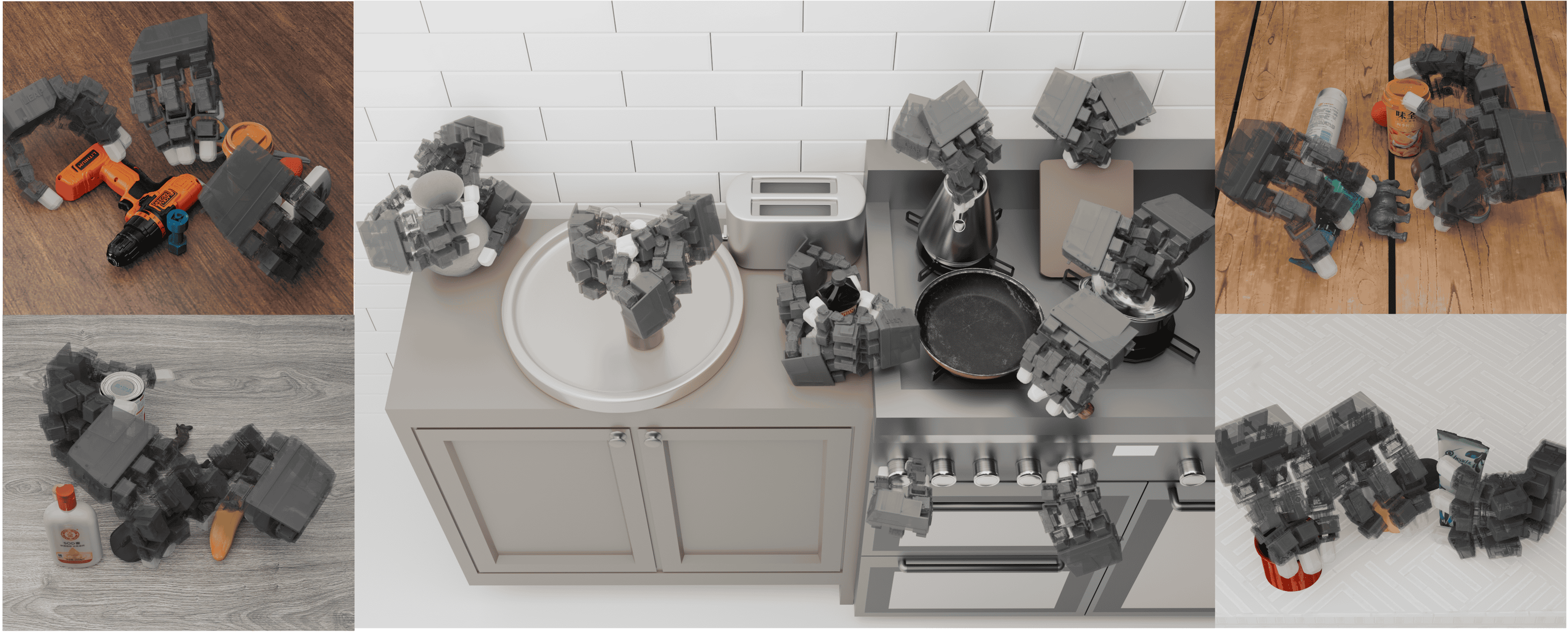}
  \centering
  \captionof{figure}{\small \our enables the generation of dexterous, functional grasps in cluttered, realistic scenes. It first uses a vision-language model to identify task-relevant regions on objects, then uses geometrically precise force closure in simulation to ground the finger joints. The resulting dataset, and the diffusion model trained on it, encode both semantic and geometric understanding of the scene without any hand-collected data.
  \label{fig:teaser}
  }
  \vspace{-0.1in}
}
\makeatother
\maketitle
\thispagestyle{empty}
\pagestyle{empty}

\begin{abstract}
Large Vision Models trained on internet-scale data have demonstrated strong capabilities in segmenting and semantically understanding object parts, even in cluttered, crowded scenes. However, while these models can direct a robot toward the general region of an object, they lack the geometric understanding required to precisely control dexterous robotic hands for 3D grasping. To overcome this, our key insight is to leverage simulation with a force-closure grasping generation pipeline that understands local geometries of the hand and object in the scene. Because this pipeline is slow and requires ground-truth observations, the resulting data is distilled into a diffusion model that operates in real-time on camera point clouds. By combining the global semantic understanding of internet-scale models with the geometric precision of a simulation-based locally-aware force-closure, \our achieves high-performance semantic grasping without any manually collected training data.  For visualizations of this please visit our website at \url{https://ifgrasping.github.io/}
\end{abstract}

\renewcommand\thefootnote{\fnsymbol{footnote}}
\footnotetext[1]{These authors contributed equally.\\
Correspondence to: muxinl@andrew.cmu.edu and kshaw2@andrew.cmu.edu}

\setcounter{figure}{1}
\begin{figure*}[!t]
 \centering
 \includegraphics[width=\linewidth]{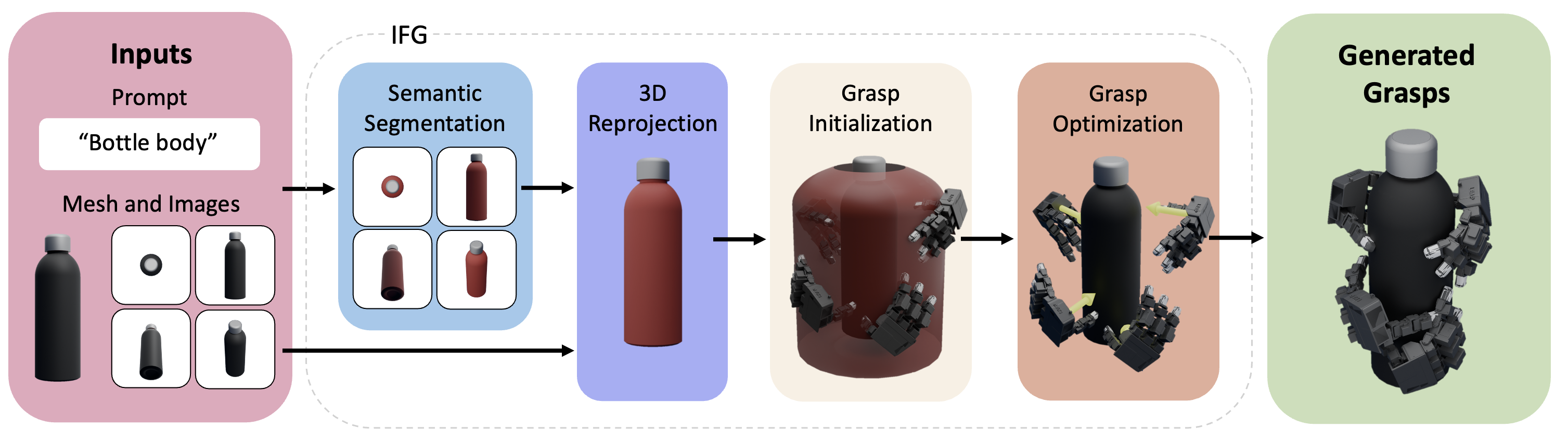}
 \vspace{-0.15in}
 $ $\caption{\small{\our takes an object mesh and a task prompt as input. To incorporate semantic understanding, it renders the object from multiple viewpoints, applies a VLM-based segmentation model combining SAM\cite{kirillov2023segment} and VLPart\cite{sun2023goingdenseropenvocabularysegmentation}, and reprojects the results into 3D space to identify task-relevant regions. For geometric grounding, it initializes a force closure objective at these regions and optimizes for functional grasps. The resulting data is then used to train a diffusion model for fast grasp synthesis from depth.}}
\vspace{-0.15in}
 \label{fig:flow_chart}
\end{figure*}

\section{INTRODUCTION}
Recent advances in vision-language models (VLMs) have led to impressive results across a range of perception tasks, including image captioning, visual question answering, and open-world object recognition. Trained on large-scale datasets pairing images with natural language, these models exhibit a strong ability to align visual and linguistic information, enabling semantic understanding that generalizes across diverse contexts. This success has inspired interest in leveraging VLMs for robotics applications such as instruction following, semantic goal specification, and high-level planning.

While these initial applications show promise, significant limitations remain. Most notably, current VLMs lack a grounded understanding of physical space—they cannot reason about 3D geometry, spatial relationships, or the dynamics of physical interaction. Consequently, they struggle with planning or executing precise motor actions in the real world. Although VLMs can identify and describe visual content, they do not inherently understand how to interact with it. This disconnect between perception and control poses a major challenge in robotic grasping systems.

We seek an approach that avoids manual data collection methods like teleoperation while enabling this geometric understanding. A promising direction involves synthetic grasp generation frameworks, which produce large datasets of grasp poses through an optimization process guided by energy functions that approximate force closure, along with evaluation pipelines in simulation. These datasets are often used to train diffusion-based grasp samplers. However, a significant portion of the generated grasps are physically implausible or unnatural. Because grasp proposals are initialized by sampling points around the object’s convex hull, many grasps target physically inaccessible or unsuitable regions.

Moreover, downstream manipulation tasks require the hand to interact with specific, task-relevant regions of objects such as a handle or button. Existing synthetic grasping pipelines generate grasps indiscriminately over the object surface, leading  datasets that are poorly aligned with the needs of task-conditioned manipulation.

Our approach addresses this gap by combining the high-level semantic understanding of VLMs with physically grounded, task-aware synthetic grasping generation. To this end, we propose a pipeline that first translates semantic input specifying a task into predictions of useful regions on objects using a VLM. Then, we seed the grasp generation process with this prior to enable semantic-guided grasp synthesis, producing stable, natural grasps aligned with the demands of the task. Our pipeline is highly parallelizable, efficient, and compatible with arbitrary objects, scenes (including cluttered environments), and dexterous hands. This pipeline generates semantically meaningful grasps without any teleoperation or video data.

 \begin{figure*}[t]
 \centering
 \includegraphics[width=\linewidth]{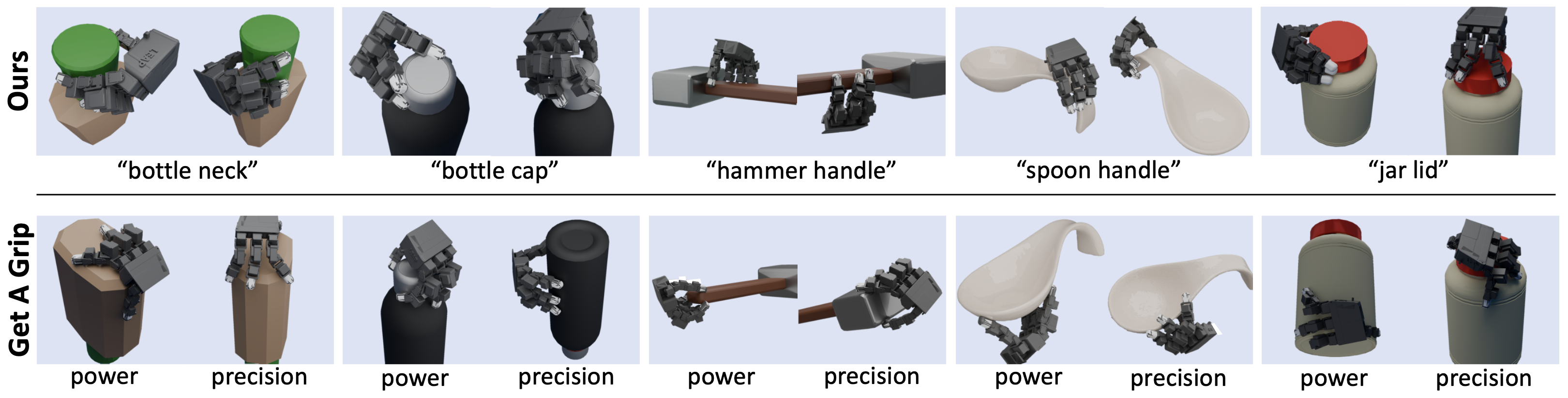}
\caption{\small{Compared to Get a Grip's synthetic grasp generation method, our method produces more human-like grasps. For instance, Get a Grip often grasp on the bottom of the bottle, while our method knows to robustly grasp the neck.  Please see our \href{https://ifgrasping.github.io/}{website} for 3D visualizations.}}
\vspace{-0.2in}
 \label{fig:qualitative_comparison}
\end{figure*}

\begin{figure}[!b]
 \centering
 \includegraphics[width=\linewidth]{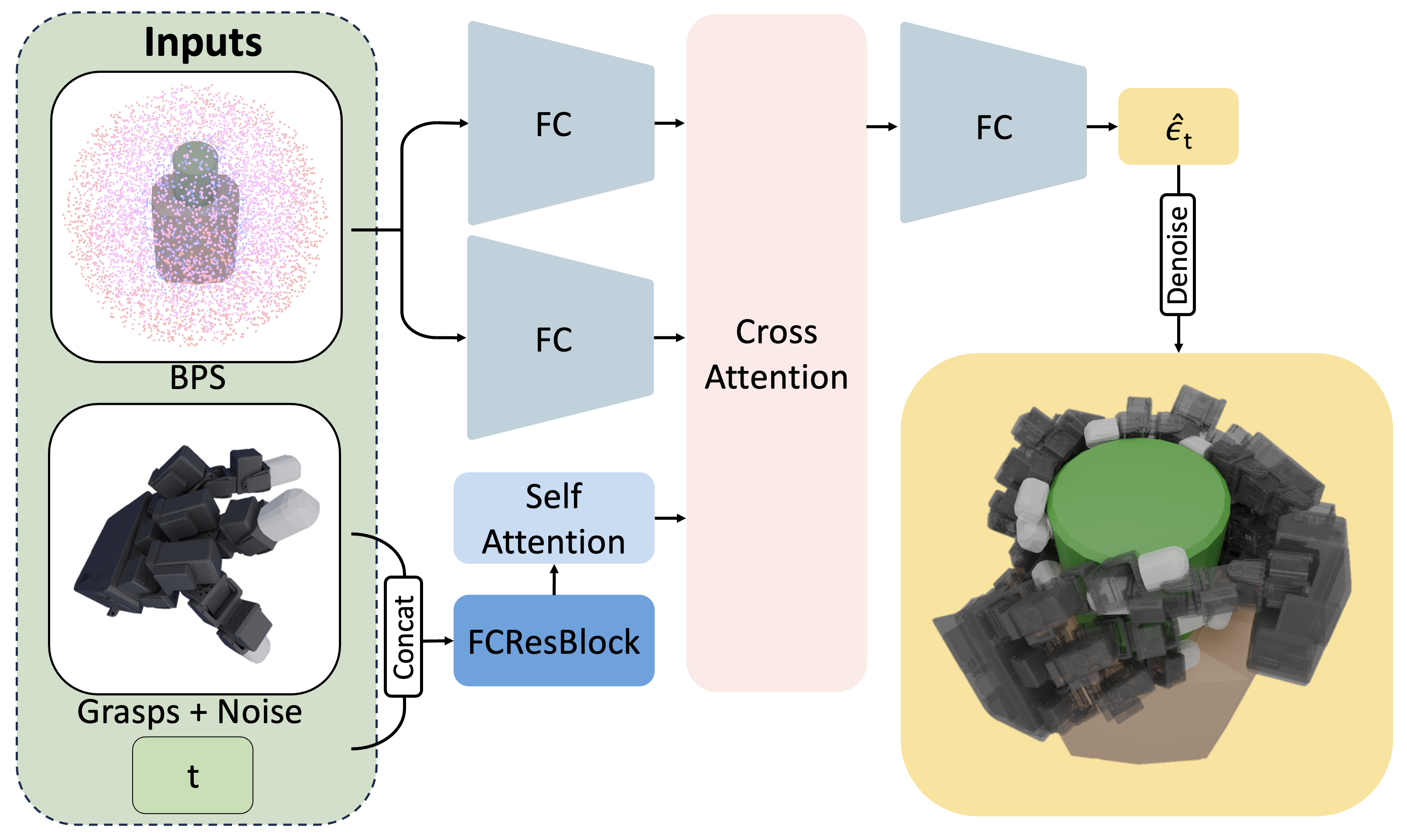}
\caption{\small{To enable real-world deployment, the generated grasp data is distilled into a diffusion model. This model is conditioned on a Basis Point Set (BPS) computed from depth camera data, along with a noisy grasp input. Through the denoising process, the model produces refined grasps on the object. The architecture of the diffusion model follows a similar design to DexDiffuser \cite{weng2024dexdiffuser}.}}
\vspace{-0in}
 \label{fig:diffusion}
\end{figure}

\section{RELATED WORKS}

\subsection{Dexterous Grasping Generation in Simulation}
GraspIt! adapts Eigengrasps from a manual database for fast grasp planning, but it doesn't generalize to any object category \cite{miller2004graspit}.  Similarly, \cite{chen2022learning} also uses shape augmentations. A line of works optimize for grasps contacts using differentiable simulation \cite{turpin2023fastgraspd, turpin2022grasp, turpin2023fast, attarian2023geometry} or by matching hand-object geometry \cite{attarian2023geometry}. DexGraspNet and its follow-up initialize the hand poses in the convex hull of the object and then optimize using the force closure objective \cite{wang2023dexgraspnet, zhang2024dexgraspnet}. We use this objective function to generate grasps as well in \our. Concurrently developed with our work, \cite{he2025dexvlgdexterousvisionlanguagegraspmodel} uses VLMs \citep{liu2023partsliplowshotsegmentation3d, tang2025segmentmesh} to produce semantic regions for grasp generation.

Many grasping pipelines use the data generated to train neural network models.  These models can aid in faster generation, the removal of privileged information, or enable generalization.  Many use a VAE model \cite{kingma2013auto} to enable this faster generation \cite{zhang2024dexgraspnet, li2023gendexgrasp,fang2020graspnet, jiang2021hand, lum2024get}.  Other newer works use the more powerful diffusion model \cite{ho2020denoising} to improve results. \cite{weng2024dexdiffuser} To remove the reliance on privileged geometry, point-cloud conditioning with Pointnet \cite{qi2017pointnet,qi2017pointnet++} are used \cite{wang2021graspness}. A parallel technique is NeRF \cite{lum2024get}.  Finally, some use quick test time adaptation to improve the grasping quality past the learned model \cite{xu2023unidexgrasp, li2023gendexgrasp}.

\subsection{Vision-based Dexterous Grasping}
Instead of generating grasps in simulation there are many datasets that have human hand and object interaction in them.  Some datasets have ground truth data using motion capture devices \cite{tao2025dexwild, shaw2024bimanual} but are more limited in size and variety of grasps \cite{chao2021dexycb, banerjee2024introducing}. Adding contact information between the object and the hand can help with fine-grained control \cite{fan2023arctic, brahmbhatt2019contactdb, brahmbhatt2020contactpose}. While there are larger datasets available, \cite{grauman2022ego4d, damen2020epic}, they do not have ground truth hand and object poses. This means that vision methods must be used to extract this, which works in varying accuracy \cite{rong2020frankmocap, ye2022s, ye2023diffusion}. Once extracted, these grasps can be used in robot hand systems.

A wide range of recent studies have tapped into large-scale human activity datasets to improve various aspects of robot learning. Some focus on deriving cost functions from human behavior \citep{shao2021concept2robot, chen2021dvd, bahl2022human, ma2022vip}, while others map human and robot actions to one another \cite{sivakumar2022robotic, handa2020dexpilot, zhang2023handypriors}, whether through aligned demonstrations \citep{sharma2019third, shaw2023videodex, patel2022learning}, unaligned examples \citep{smith2019avid}, or direct action correspondences \citep{schmeckpeper2020reinforcement}. In addition, the inherent structure of certain datasets—such as those involving tool use \citep{young2020visual}, or temporal sequences of hand-object interactions \citep{lee2017learning,kannan2023deft}—has been used to infer actions or detect salient features like keypoints.  

\subsection{VLMs for Robotic Grasping}
Recent works have explored integrating large-scale models with robotic grasping, particularly for two-finger grippers. For example, \cite{huang2023voxposer} and \cite{sharma2022correcting} extract affordances and constraints from LLMs and VLMs to build 3D value maps, which are then used by motion planners to synthesize trajectories in a zero-shot manner. Similarly, \cite{agarwal2023dexterous} incorporates large-scale models but relies on simulation to train downstream policies. Other approaches \cite{nair2022learning, yu2023language, liang2022code} generate vision-language-action (VLA) representations or language-based plans that can be executed on robotic systems.
\vspace{0.2in}

\section{METHOD}
The goal of \our is to learn a general-purpose dexterous grasping affordance model that takes as input a scene point cloud and a text prompt specifying the object to grasp, and outputs a feasible grasp for a robot hand. To enable this, we must first generate a large grasping dataset with geometrically accurate and semantically meaningful grasps. As shown in Figure \ref{fig:flow_chart}, given an object mesh and a task prompt, our data generation pipeline identifies task-relevant regions by rendering the object from multiple viewpoints, applying a VLM-based segmentation model, and reprojecting the results into 3D. These semantic regions then guide a grasp optimization process that enforces both stability and functional relevance. The resulting diverse set of robust grasps is distilled into a diffusion model that predicts executable grasps directly from depth input, enabling fast and deployable grasp synthesis in real-world scenarios. We present the pseudocode of our method in Algorithm \ref{alg:ifg}.

\subsection{Dexterous Grasp Formulation}
We formulate dexterous grasps as follows. A dexterous grasp $g$ is defined as $g=(T, R, \theta)$, where $T\in \mathbb{R}^3$ and $R\in SO(3)$ represent the translation and rotation of the wrist pose, and $\theta\in\mathbb{R}^{\text{DoF}}$ denotes the joint angles of the hand ($\text{DoF}=16$ for LEAP Hand \cite{shaw2023leaplowcostefficient}). 

\subsection{Useful Region Proposal}
IFG leverages knowledge from a VLM $f$ to identify objects of interest and part-level useful regions. To extract 2D semantic knowledge to 3D scenes, we also use a language-conditioned segmentation model $g$ to isolate the object and a part-level segmentation model $h$ to identify regions of interest on the object. Given an object mesh $O$ with $k$ faces $F=\{f^{(1)},...,f_{(k)}\}$, we take a set of $n$ RGB images $V=\{v^{(1)},…, v^{(n)}\}$ from angles uniformly sampled on a camera initialization surface $S$. For single object settings $S$ is a spherical surface, and for clustered scenes it is the dome segmented from a sphere to avoid visual occlusion. A VLM is prompted with $V$ to produce $m$ semantic labels of useful regions of $O$ denoted as $f(V)=R=\{r_1,…, r_m\}$. For each label $p_i\in P$, $g$ and $h$ together produces part-level segmentation masks for each image in $V$ conditioned on $r_i$, represented as $h\circ g(r_i)=S_i=\{s_i^{(1)},…, s_i^{(n)}\}$, where $s_i^{(j)}$ segments the regions of $v^{(m)}$ that belong to $r_i$. We use SAM \cite{kirillov2023segment} as the object segmentation model and VLPart \cite{sun2023goingdenseropenvocabularysegmentation} as the part-level segmentation model.

The 2D segmentation masks $S_i$ are deprojected back to 3D points on the object mesh $P_i=\{p_i^{(1)},…, p_i^{(n)}\}$. However, from certain camera angles a part may be occluded, leading to incorrect segmentation. To address this, $P_i$ further undergoes heuristic-based filtering. 

A two-means based clustering process assigns each $p_i$ to one of two groups based on the size of its segmentation mask $s_i$ in terms of masked pixel number to filter out unsuccessful segmentation masks. The larger group from the clustering process $\hat{P_i}$ is considered the valid deprojected points. Each point is then associated with the closest face on the object mesh to produce a tally of face counts $T_i=\{t_i^{(1)}, ..., t_i^{(k)}\}$, where $t_i^{(j)}\in\mathbb{N}$. A voting algorithm selects the 60\% top faces of the object mesh as the useful region of the object used for the next stage, which we refer to as $U$. 

\subsection{Geometric Grasp Synthesis}
We compute the segmented convex hull of the object to include only faces projected from $U$. For each grasp, the hand is initialized on the inflated convex hull by farthest point sampling with random noise added to the pose and finger joint angles. An optimization process performs gradient descent against an energy term
$$
E=E_{\text{fc}}+w_{\text{dis}}E_{\text{dis}}+w_{\text{joints}}E_{\text{joints}}+w_{\text{pen}}E_{\text{pen}}+w_{\text{spen}}E_{\text{spen}}
$$
where $E_{\text{fc}}$ approximates force closure of the grasp, $E_{\text{dis}}$ encourages hand-object proximity, based on the contact points of the hand, $E_{\text{joints}}$, $E_{\text{pen}}$, and $E_{\text{spen}}$ respectively penalizes joint violations, hand-object penetration, and self-penetration of the hand. For the single object setting, we exclude the tabletop by setting $w_{\text{spen}}=0$ to produce more diverse grasps. This pipeline is similar to Get a Grip's synthetic pipeline except for a few key modifications. Instead of using precision grasps, which sample contact points only on the fingertips of the hand, we use power grasps by sampling over the inside regions of all fingers, which produce more stable grasps and thus yield a higher success rate. For each grasp, we initialize hand positions on the segmented convex hull instead of the entire hull.

\subsection{Simulation Evaluation}
To ensure the robustness of generated grasps, we perform tasks with them in a simulation environment. Each evaluation proceeds in three phases: (1) the grasp and object are initialized in a simulation environment, (2) fingers are closed to secure the object, and (3) task execution is performed.  Following Get a Grip, we use a smooth label for each grasp by applying slight perturbations on the finger joint angles to produce $d$ associating grasps, all of which are evaluated in simulation. The hard success rates of all $d+1$ grasps are averaged to produce the smooth label for the grasp. Grasps with low success rates are filtered out to produce a dataset $G$ of robust, force-closure power grasps. For our experiments, $d=5$. 

\begin{algorithm}[t]
\caption{IFG}
\label{alg:ifg}
\begin{algorithmic}[1]

\Require VLM $f$, segmentation models $g,h$, object mesh $O$ with faces $F=\{f^{(1)},\dots,f^{(k)}\}$, $n$ views $V=\{v^{(1)},\dots,v^{(n)}\}$ from camera surface $S$

\Statex \textbf{Semantic Segmentation Region Extraction}
\State Query VLM: $R = f(V) = \{r_1,\dots,r_m\}$ semantic labels
\For{each label $r_i \in R$}
    \State Obtain masks $S_i = h \circ g(r_i) = \{s_i^{(1)},...,s_i^{(n)}\}$
    \State Deproject masks: $P_i = \{p_i^{(1)},...,p_i^{(n)}\}$
    \State Filter $P_i$ with two-means clustering by mask size
    \State Map filtered points $\hat{P_i}$ to nearest faces, tally counts $T_i$
\EndFor
\State Select top 60\% faces $\; U \subseteq F$ as useful regions

\Statex \textbf{Geometric Grasp Synthesis}
\State Build convex hull from $U$; inflate for sampling
\For{each grasp initialization}
    \State Place hand on hull via farthest point sampling $+$ random noise
    \State Optimize energy 
    \[
      E = E_{\text{fc}} + w_{\text{dis}}E_{\text{dis}} + w_{\text{joints}}E_{\text{joints}} 
          + w_{\text{pen}}E_{\text{pen}} + w_{\text{spen}}E_{\text{spen}}
    \]
    \State Obtain candidate grasp
\EndFor

\Statex \textbf{Simulation Evaluation}
\For{each grasp}
    \State Generate $d$ perturbed grasps by varying joint angles
    \State Simulate Lift / Pick \& Shake tasks in IsaacGym
    \State Assign smooth label as mean success over $d+1$ trials
\EndFor
\State Filter grasps with low success $\rightarrow G$
\end{algorithmic}
\end{algorithm}

\subsection{Diffusion Model Distillation}

While the grasping pipeline can generate numerous candidate grasps from an object mesh, it is not directly deployable in real-world scenarios due to practical constraints.  The generation process is quite slow, object mesh is not readily available, and the generation process often does not always return successful grasps.  
Inspired by \cite{weng2024dexdiffuser}, our diffusion model takes as input a Basis Point Set (BPS) which is a structured point cloud that can be readily obtained from the object mesh using a depth camera \cite{prokudin2019efficient}. Additionally, the model receives a noisy grasp hypothesis. Through the denoising process, the diffusion model refines this noisy input into a feasible and executable grasp.  This downstream diffusion model inherits both the geometric reasoning capabilities of the training pipeline and the semantic understanding provided by the vision-language model (VLM), as illustrated in Figure \ref{fig:diffusion}.

\begin{table}[t]
\centering
\begin{tabular}{@{}lcc@{}}
\toprule
\textbf{Object} & \textsc{Get a Grip} & \textsc{Ours} \\
\midrule
water bottle            & 49.1 & \textbf{62.8} \\
large detergent bottle  & 51.2 & \textbf{62.5} \\
spray bottle            & 43.1 & \textbf{54.5} \\
pan                     & 48.1 & \textbf{52.1} \\
small lamp              & 56.8 & \textbf{85.7} \\
spoon                   & 42.7 & \textbf{50.9} \\
vase                    & 32.2 & \textbf{55.9} \\
hammer                  & 45.8 & \textbf{45.8} \\
shark plushy            & 19.8 & \textbf{25.1} \\
\bottomrule
\end{tabular}
\caption{A selection of individual success rates out of the 35 objects we generate on in single-object scenes.  Ours generation outperforms the baseline Get a Grip~\cite{lum2024get} due to improved grasp initializations from the VLM.}
\label{tab:single_object_eval}
\vspace{-0.2in}
\end{table}

\section{Experimental Setup}
Datasets of grasps are generated on diverse objects in both single-object and clustered-scene settings, followed by extensive simulation to evaluate robustness. The evaluation addresses four key questions: (1) Can robust and stable grasps be produced on individual objects? (2) In clustered scenes, can the object of interest be identified and grasped without collision? (3) Do the resulting grasps exhibit natural, human-like qualities suitable for functional manipulation? (4) To what extent does semantic, part-level conditioning via segmentation improve grasp robustness and naturalness?

\textbf{Task Setup.} Two evaluation settings are considered. In the single-object case, 24 diverse objects from Get a Grip’s dataset are used; each object is sampled at 5 scales, with 200 grasps generated by both our method and the baseline. For clustered scenes, 35 dense scenes from DexGraspNet2 are selected. Each scene contains on average 3–4 objects, with 3–4 segmentation prompts per object, and 200 grasps generated for each prompt–object pair. Baselines sample 256 grasps per scene. All objects are drawn from common daily manipulation tasks, and all grasps are executed using the LEAP Hand \cite{shaw2023leaplowcostefficient}. Finally, a diffusion model is trained to verify that the generated grasps can be distilled into a policy operating directly on proprioceptive data obtainable in the real-world.  Please also see our website at \url{https://ifgrasping.github.io/} for more visualizations of these results.

\begin{table}[!b]
\centering
\begin{tabular}{@{}lcc@{}}
\toprule
Method                        & Pick \& Shake (\%) & Lift  (\%) \\
\midrule
Ours                          & 16.14            &     51.11              \\
Get a Grip       & 11.82            &      50.93             \\
\bottomrule
\end{tabular}
\caption{Single-object grasp generation evaluation in Isaac Gym. Our method outperforms Get a Grip by leveraging VLM-based part-level awareness. Successful grasps are filtered and used to train the diffusion model.}
\label{tab:mode_success_rates_single}
\vspace{-0.2in}
\end{table}

\begin{figure}[!t]
 \centering
 \includegraphics[width=\linewidth]{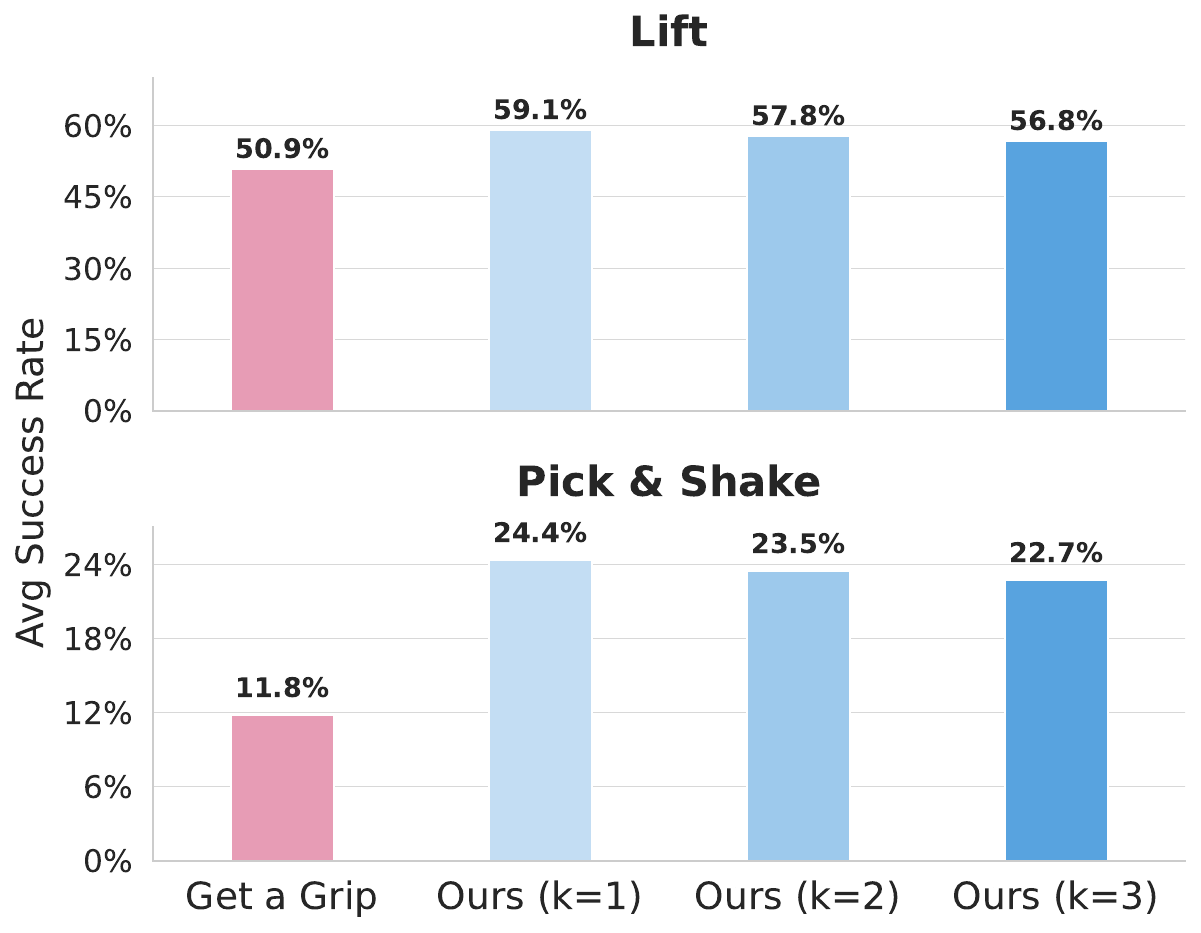}
\caption{Single Object evaluation in the Lift and Pick and Shake Task.  Ours outperforms on the top three segmentation prompts compared to the Get a Grip baseline generation process due to the guidance that the prompt and the VLM provide on the grasping generation process.}
\vspace{-0.1in}
 \label{fig:single_topk_success}
\end{figure}

\begin{table}[t]
\centering
\begin{tabular}{@{}lcc@{}}
\toprule
Method                        & Lift (\%) \\
\midrule
Ours            &    32.23               \\
GraspTTA        &    25.64                           \\
ISAGrasp        &    32.51                          \\
DexGraspNet2        & 36.71                            \\
\bottomrule
\end{tabular}
\caption{\our can generate grasps with similar lift success rates to baseline models trained on preprocessed and filtered DexGraspNet2's Dataset, showing our strong grasp generation capabilities.}
\label{tab:mode_success_rates_crowded}
\vspace{-0.2in}
\end{table}

\textbf{Simulation Evaluation.} We test our grasps in IsaacGym \cite{makoviychuk2021isaac}. For the single-object setting, two tasks are designed: Lift, which raises the wrist vertically to test grasp firmness, and Pick \& Shake, which lifts the object slightly and applies perturbations to the wrist. A task is considered successful if the object’s relative pose to the palm remains stable throughout execution. Collision checking is enforced during the entire process. For clustered scenes, we evaluate grasps only on the Lift task, since shaking in a dense environment often leads to trivial collisions.

\section{RESULTS}

\begin{table*}[ht]
\centering
\begin{tabular}{@{}lcccc@{}}
\toprule
\textbf{Object} & \textsc{DexGraspNet2} & \textsc{GraspTTA} & \textsc{ISAGrasp} & \textsc{Ours} \\
\midrule
Tomato Soup Can         & 47.8 & 38.3 & \textbf{52.0} & 45.5 \\
Mug                     & 33.2 & 26.9 & 22.6 & \textbf{60.4} \\
Drill                   & 32.1 & 20.8 & 36.4 & \textbf{57.5} \\
Scissors                & 9.7  & 0.0  & \textbf{33.7} & 20.2 \\
Screw Driver            & 0.0  & 8.3  & \textbf{40.0} & 22.0 \\
Shampoo Bottle          & 50.6 & 25.4 & 18.8 & \textbf{53.1} \\
Elephant Figure         & 23.6 & 29.6 & 24.2 & \textbf{35.8} \\
Peach Can               & \textbf{61.8} & 28.0 & 55.3 & 60.3 \\
Face Cream Tube         & 32.1 & 22.5 & 20.7 & \textbf{35.5} \\
Tape Roll               & 22.7 & 13.9 & 9.8 & \textbf{43.2} \\
Camel Toy               & 12.8 & 14.3 & 21.3 & \textbf{21.8} \\
Body Wash               & 40.2 & 22.3 & 29.4 & \textbf{58.3} \\
\bottomrule
\end{tabular}
\caption{Grasp success rates for crowded-scene evaluation on the lift task.  The VLM enables \our to focus on objects of interest and exceeds them in performance compared to baselines \cite{zhang2024dexgraspnet,jiang2021hand, chen2022learning}}.
\label{tab:crowded_scene_eval}
\vspace{-0.3in}
\end{table*}

\subsection{Single Object Grasping}
A useful grasp is not only robust but also natural, both of which can be achieved through our method. To demonstrate this, we evaluate our method against Get a Grip \cite{lum2024get} on 35 diverse objects from their dataset used in daily scenarios. To assess robustness, grasps are evaluated in simulation under two tasks, Pick \& Shake and Lift. As illustrated in Table \ref{tab:mode_success_rates_single}, IFG achieves higher success rates on both tasks, demonstrating that conditioning grasp generation on part-level segmentation produces more robust grasps. Table \ref{tab:single_object_eval} further presents detailed success rates of a diverse set of objects from our data. Moreover, from qualitative comparisons, our grasps are more natural: they are concentrated on the object regions that humans typically interact with in real-world use, while many of Get a Grip’s grasps that pass simulation checks are not aligned with functional usage due to the absence of guidance during initialization. Shown in Figure \ref{fig:qualitative_comparison}, their grasps tend to grasp the head of a hammer since it covers a high percentage of the convex hull, while our grasps initialized on the segmented convex of the handle are functionally correct. With our method outperforming the baseline on both robustness and naturalness, we hypothesize that semantically conditioned grasps improve robustness because everyday objects are designed with affordances that support secure functional grasping, and semantic conditioning aligns grasp generation with these regions.

\subsection{Multi-object Dense Scene Grasping}
\begin{figure}[!b]
 \centering
 \includegraphics[width=\linewidth]{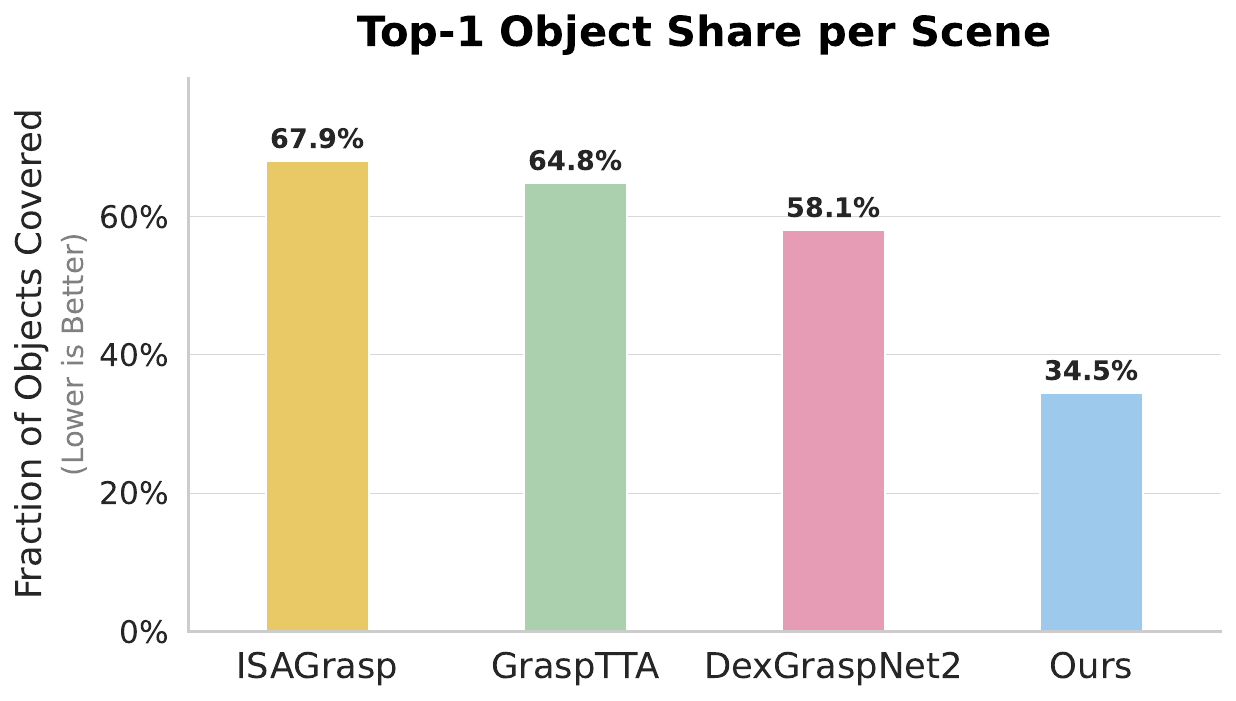}
\caption{When generating grasps, confidence-based methods generate most of their grasps on the easiest-to-grasp object. On the other hand, our method can be controlled to grasp any specific object due to segmentation conditioning.  Therefore the easiest object is grasped less often.}
\vspace{-0.1in}
 \label{fig:concentration_grasps}
\end{figure}

Daily scenarios are often not so simple as single object settings because they involve many clustered objects. A grasp proposal pipeline must therefore be able to identify the object of interest and generate firm grasps while avoiding collision with others. Get a Grip does not address multi-object scenes, so we compare against the crowded scene grasp generation models in DexGraspNet2 \cite{zhang2024dexgraspnet}. DexGraspNet2 retargets GraspNet-1Billion data \cite{fang2020graspnet} into a diffusion model and adapts several single-object networks as baselines. Their approach ranks points on the scene point cloud with an MLP to propose grasp seeds, but cannot control which object is grasped. In addition, their ranking method tends to be biased toward easy targets, as shown in Figure \ref{fig:concentration_grasps}. In contrast, our method selects via semantic segmentation prompts and avoids overfitting to easy-to-grasp regions. We evaluate IFG on clustered, dense scenes with harder objects from DexGraspNet2. Figure \ref{fig:teaser} shows our grasps on four scenes on the sides. Impressively, our synthetic generation method achieves a similar success rate compared to the baseline models distilled from preprocessed and filtered data, which is shown in Table \ref{tab:mode_success_rates_crowded}. A more detailed analysis done on individual objects across scenes is shown in Table \ref{tab:crowded_scene_eval}.

\begin{figure}[!b]
 \centering
 \includegraphics[width=\linewidth]{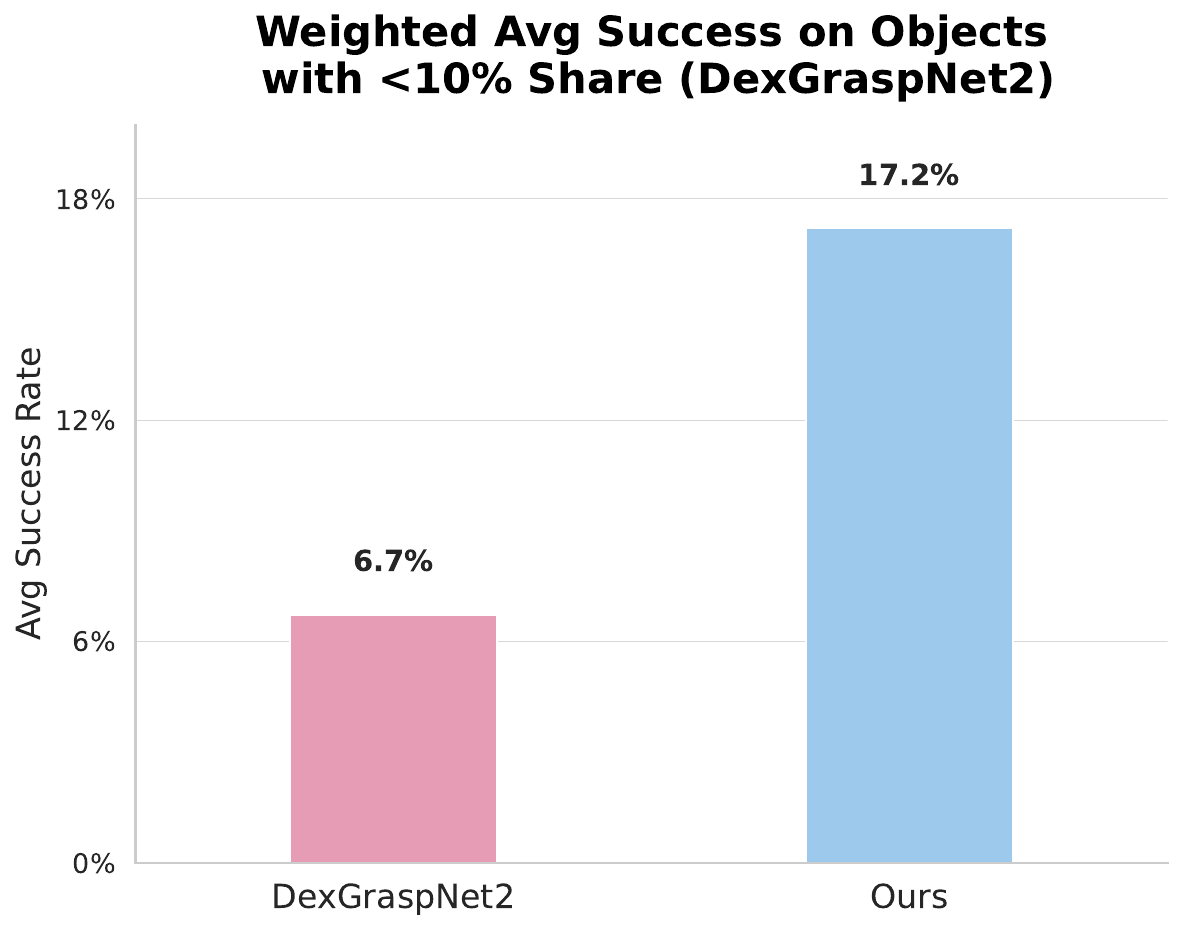}
\caption{DexGraspNet2's grasp generation model avoids hard-to-grasp objects. Our method concentrates more on these objects and achieves a higher success rate due to functional guidance from VLM-based segmentation.}

\vspace{-0.1in}
 \label{fig:least_concentrated_success}
\end{figure}

The differences between ours and DexGraspNet2's reported performance is due to two reasons. (1) both methods lift objects by 20 cm, but DexGraspNet2 counts a grasp as successful if the object rises just 3 cm, even if it slips onto nearby objects. (2) More comprehensive testing: DexGraspNet2 reports only the top-confidence grasp per scene, usually on easy-to-grasp objects on the peripheral of the scene. We evaluate over 200 grasps per scene for their baselines. As shown in Figure \ref{fig:least_concentrated_success}, on harder, we outperforms them on occluded, harder-to-grasp objects that they grasp less frequently.

\subsection{Grasp Generative Model}
The method is modular, enabling plug-and-play replacement of both the segmentation and generation modules. For grasp generation, an attention-based conditional diffusion transformer (DiT) is trained to produce grasps conditioned on the object’s BPS \cite{prokudin2019efficient} representation computed from its point cloud, following an architecture similar to that used in Get a Grip. Grasps generated by this model, trained on semantically meaningful data, are compared against those produced by Get a Grip to highlight the benefits of semantic conditioning. (The model is trained on a single object at one scale, a bottle.)

\section{Conclusion and Limitations}
We introduced \our, a pipeline that combines the semantic understanding of vision-language models with the geometric precision of force closure grasping to generate functional and robust dexterous grasps. IFG harnesses internet-scale models to identify task-relevant object regions from visual information, uses them as semantic conditioning for energy based force closure optimization, and leverages simulation evaluation as a metric for robustness. As a result, IFG produces more natural and effective grasps than prior methods, particularly in cluttered environments. The resulting data is distilled into a diffusion model, enabling real-time grasp prediction from camera input without relying on hand-collected data.

Nonetheless, our work has limitations. First, our method does not account for dynamic objects since our segmentation is performed on images from a single timestep. Potential work can be done on extending our semantic segmentation pipeline for continuous video streaming. Additionally, our method is not suited for scenarios where non-force closure grasps are required. There is still much to be done in optimization based dexterous grasp generation.

\section{Acknowledgements}
We thank Jason Liu, Andrew Wang, Yulong Li, Jiahui (Jim) Yang, Sri Anumakonda for helpful discussions and feedback. This work was
supported in part by the Air Force Office of Scientific Research (AFOSR) under Grant No. FA955023-1-0747 and by the Office of Naval Research (ONR) MURI under Grant No. N00014-24-1-2748.

\section{Contributions}
\noindent \textbf{Ray Muxin Liu} implemented the grasp-region prompting and 3D deprojection scheme, the grasp optimization module, as well as the clutter-scene grasping experiments and benchmarking in simulation, and led the manuscript writing.

\noindent\textbf{Mingxuan Li} developed the early-stage components, including the grasp-generation framework, the grasp-diffusion model, and the 2D segmentation of grasp regions. Upon graduating, Mingxuan Li concluded active contributions to the project

\noindent\textbf{Kenneth Shaw} originated the core idea and guided the research direction.

\noindent\textbf{Deepak Pathak} supervised the project.
\bibliographystyle{IEEEtran}
\bibliography{IEEEabrv,main}

\end{document}